\documentclass[journal,twoside,web]{ieeecolor}
\usepackage{jsen}
\usepackage{cite}
\usepackage{amsmath,amssymb,amsfonts}
\usepackage{algorithmic}
\usepackage{graphicx}
\usepackage{hyperref}
\usepackage{tabularx}
\usepackage{textcomp}
\usepackage{wrapfig}
\usepackage[final]{pdfpages}
\def\BibTeX{{\rm B\kern-.05em{\sc i\kern-.025em b}\kern-.08em
    T\kern-.1667em\lower.7ex\hbox{E}\kern-.125emX}}
\definecolor{abstractbg}{rgb}{0.89804,0.94510,0.83137}
\begin{document}
\null%
\includepdf[pages=-,pagecommand={},width=\textwidth]{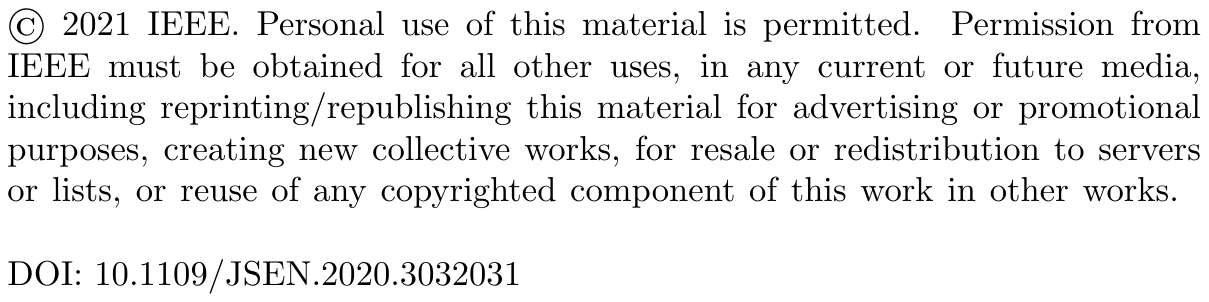}
\title{Improving Automated Sonar Video Analysis to Notify About Jellyfish Blooms}
\author{Artjoms Gorpincenko, Geoffrey French, Peter Knight, Mike Challiss, and Michal Mackiewicz
\thanks{This work was jointly supported by Cefas, Cefas Technology Limited, EDF Energy, Innovate UK under Grant 102072, Natural Environmental Research Council under Grant NE/RO12156/1 and Nvidia Corporation.}
\thanks{Artjoms Gorpincenko, Geoffrey French, and Michal Mackiewicz are with the School of Computing Sciences, University of East Anglia, Norwich NR4 7TJ, U.K (e-mail: a.gorpincenko@uea.ac.uk).}
\thanks{Peter Knight and Mike Challiss are with Cefas Technology Ltd., Cefas, Lowestoft NR33 0HT.}}

\IEEEtitleabstractindextext{%
\fcolorbox{abstractbg}{abstractbg}{%
\begin{minipage}{\textwidth}%
\begin{wrapfigure}[8]{r}{4.8in}%
\includegraphics[width=4.6in]{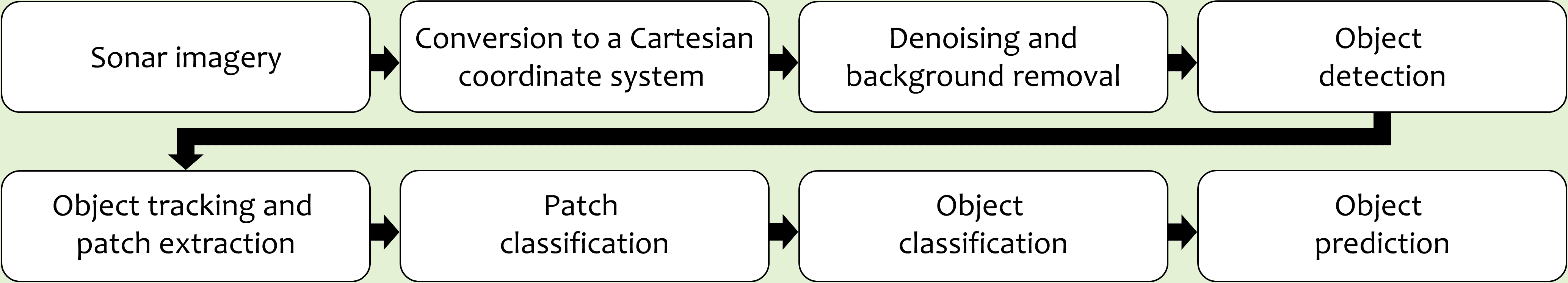}%
\end{wrapfigure}%
\begin{abstract}
Human enterprise often suffers from direct negative effects caused by jellyfish blooms. The investigation of a prior jellyfish monitoring system showed that it was unable to reliably perform in a cross validation setting, i.e. in new underwater environments. In this paper, a number of enhancements are proposed to the part of the system that is responsible for object classification. First, the training set is augmented by adding synthetic data, making the deep learning classifier able to generalise better. Then, the framework is enhanced by employing a new second stage model, which analyzes the outputs of the first network to make the final prediction. Finally, weighted loss and confidence threshold are added to balance out true and false positives. With all the upgrades in place, the system can correctly classify 30.16\% (comparing to the initial 11.52\%) of all spotted jellyfish, keep the amount of false positives as low as 0.91\% (comparing to the initial 2.26\%)  and operate in real-time within the computational constraints of an autonomous embedded platform.
\end{abstract}

\begin{IEEEkeywords}
Deep learning, jellyfish quantification, object classification, sonar imagery, underwater monitoring.
\end{IEEEkeywords}
\end{minipage}}}

\maketitle

\section{Introduction}
\label{sec:introduction}
\IEEEPARstart{J}{ellyfish} population has increased worldwide due to climate change and human activity, such as overfishing, aquaculture, and coastal constructions, that create satisfactory environments for polyp settlement and winter survival \cite{Mills2001, Pauly2009}. This phenomenon is not limited to a set of particular species and has been noticed in many different coastal ecosystems \cite{Graham2001, Link2006, Malej2012, Shiganova1998}. Moreover, this growth in numbers is unstable and often unpredictable; thanks to the explosive nature of gelatinous zooplankton life cycles, they usually come in periodic blooms \cite{Mills2001, Boero2016}. Such high volumes of jellyfish are known for clogging cooling water intakes at coastal power and desalination plants, filling and splitting fishing nets, killing captured fish, swarming beaches, and stopping ship engines \cite{Masilamani2000, Kim2016, Purcell2007AnthropogenicCO}, which incurs large damage and associated costs to a number of marine-related industries, such as travel, shipping, energy generation, and tourism. 

Imaging sonars are a good solution to the problem, as they can produce images at movie-like frame rates in the underwater marine environment, where low-light video systems struggle to provide useful imagery and notify about blooms. With recent advancements in computer vision and deep learning, it is possible to collect, process and analyse sonar data in an unmanned real-time manner. All of the above is performed by JellyMonitor \cite{8652268} - a self-contained system that is designed to monitor coastal environments and notify about the presence of several objects and phenomena, jellyfish being one of them. In this paper, we outline several issues that are present in the classification part of the project and propose a series of enhancements that are aimed to solve these problems. We empirically prove that our solution is more robust to false detections, generalises better to various types of underwater environments, and does not add much computational overhead.

The major contributions of this paper are summarised as follows:
\begin{enumerate}
    \item Robust detection and classification of underwater objects, in an unmanned real-time manner;
    \item Application of generative adversarial networks to sonar imagery;
    \item A two-step classification framework that allows for fast, independent training and better results at test time.
\end{enumerate}
To our knowledge, JellyMonitor is the only system which can reliably spot and report jellyfish blooms, as well as self-sustain underwater for up to three months.

\section{Related Work}
Similar systems mostly use traditional cameras to detect jellyfish \cite{martin2020jellytoring}. Whereas optical instruments can provide high resolution RGB images and are effective at short distances, they are unable to produce good underwater imagery at deep water levels and at night without sufficient illumination. There were attempts to capture data with cameras being above the water level, i.e., mounted on a wharf \cite{Llewellyn2016, 7379218} and built into an unmanned aerial vehicle \cite{7358813, 7381601}. In the first case, two 100W floodlights were required to brighten the area of interest, while data was gathered from very shallow waters (5m - 10m); the latter could analyse the water surface only. It is also unknown how robust these approaches are to reflections that occur in sunny weather. A number of studies attempted underwater imagery collection and analysis \cite{4099033, 7358846, 1255509, 0f58b71b156144388d7e97020b126d2f, 1315079}, however, did not gather enough data to adequately evaluate the performance, hence it is unknown how they would process long-term observations. Some systems did not consider real life conditions and were tested on synthetic data \cite{Matsuura2007} or in manually set up environments where the presence of noise is minimal e.g., well lit tanks \cite{7761078}. The idea of slicing jellyfish by cutting nets or blades showed its effectiveness \cite{Kim2016, Park2011}; we deviate from that approach as upon being attacked, medusae release large amounts of eggs and sperm that later form planulas, thus amplifying population increase in the future.

Deep learning has achieved state of the art results in various computer vision tasks, object classification \cite{Huang2017, Simonyan2014, Szegedy2015} being one of them. As it offers high accuracy and speed during test time and learns data distributions in an unmanned manner, it became the preferred option for many applications where enough computational power is available. Convolutional neural networks (CNNs) were applied to underwater sonar imagery and showed good results \cite{nguyen2020study, 2018mst, 8984248, 04467e4819d546b788da3fb700f377b6}. Preceded by traditional methods that spot and track moving objects in captured videos, JellyMonitor uses a CNN to extract visual features and classify a variety of phenomena.  

Generative adversarial networks (GANs) \cite{DBLP:conf/nips/GoodfellowPMXWOCB14} have been improving rapidly, and can now produce highly realistic machine generated images \cite{karras2019analyzing, brock2018large}. GAN-based methods were successfully applied to sonar data \cite{TERAYAMA2019102000, LEE2019152}, however, they were both based on image-to-image translation \cite{CycleGAN2017} between acoustic and optical images. We propose a conditional StyleGAN-based \cite{karras2019style} method to generate sonar data, which does not require other visual domains, i.e. RGB footage.

\section{Data Overview}
Data collection is an ongoing process for the JellyMonitor project; it is performed by placing the imaging sonar underwater, together with a battery and a hard drive that stores captured footage. We use Nvidia Jetson TX2 as the embedded processor which has a built in graphics processing unit (GPU) that allows to execute deep neural networks offshore quickly. At the initial stages of this research, it was important to configure the range of the acoustic device, collect initial data, perform battery durability tests, and make sure that the system can sustain underwater. This was done in tank and then in a harbour, which resulted in necessary adjustments, as well as many clear images of moon jellyfish (Aurelia Aurita), as they were released manually within the sonar's field of view. Then, to obtain footage in realistic scenarios, where the presence of human actions is minimal, JellyMonitor was deployed on the seabed off the coast. The deployment was successful and brought a large amount of data (see Table \ref{datasummary}, Year 2017). Such large quantity of imagery contains millions of objects; even with automated image analysis, it would require months, if not years, to spot and label the main targets of interest, such as jellyfish, seaweed, and fish. Therefore, we changed our data collection strategy for all the next deployments - the system would be turned on only for short periods of time between tides, when tidal flow is at its fastest. Jellyfish are stolid animals, and the chances of spotting them are much higher when marine currents are present. Moreover, it allows for the battery to last longer, resulting in more observations. Spotted objects are put in 6 categories, as follows:
\begin{itemize}
    \item Background - noise that surpassed thresholds during the denoising process and caused a false detection;
    \item Jellyfish - mostly moon jellyfish, the main target of interest;
    \item Artefacts - sonar signal reflections from the water surface;
    \item Fish - a very common object in any waters;
    \item Seaweed - countless species of marine plants that can also cause damage to the industry;
    \item Sediment - air bubbles and diffuse clouds of sediment that are caused by water turbulence.
\end{itemize}
All phenomena that get detected by the object tracking algorithm are later manually labeled by a human observer. Both the algorithm and the labeling application are described in details in the previous publication \cite{8652268}.

\begin{figure}[!t]
\centerline{\includegraphics[width=\columnwidth]{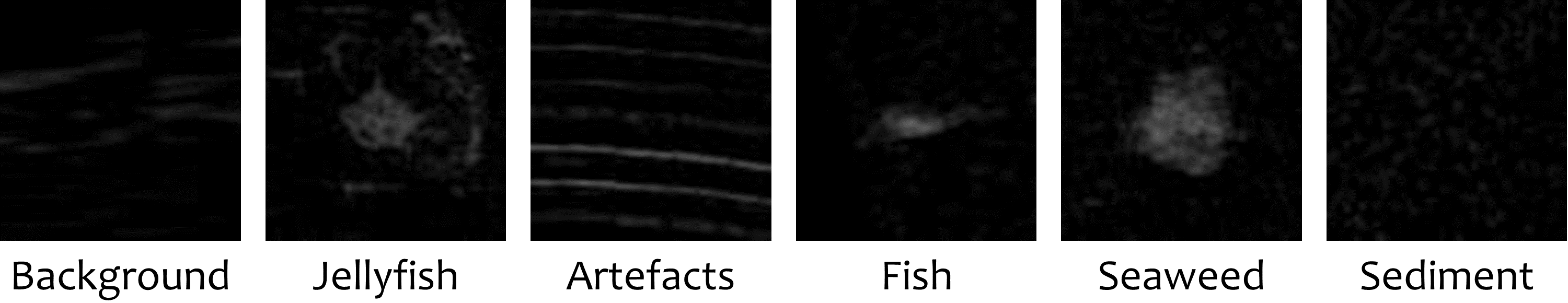}}
\caption{Example images of categories present in the datasets.}
\label{categories}
\end{figure}

\begin{table}
\caption{Summary of obtained imagery and labelled objects}
\setlength{\tabcolsep}{4pt}
\label{datasummary}
    \begin{tabular}{|l|r|r|r|r|r|r|r|}
    \hline
         Description & Frames & B & J & A & F & Sw & Sd \\ \hline \hline
         2015 Trial & 186,164 & 56 & 179 & 138 & 0 & 0 & 0\\
         2015 Trial & 180,277 & 98 & 57 & 49 & 169 & 38 & 0\\
         2016 Trial & 598,926 & 77 & 8 & 109 & 67 & 88 & 13\\
         2017 Deployment & 14,119,741 & 281 & 70 & 703 & 565 & 372 & 175\\
         2018 Deployment & 1,903,140 & 1675 & 44 & 370 & 1006 & 48 & 1670\\ 
         2019 Deployment & 2,343,344 & 3313 & 223 & 486 & 4583 & 1010 & 1278\\ \hline
         Total & 19,331,592 & 5500 & 581 & 1855 & 6390 & 1556 & 3136 \\
    \hline
    \multicolumn{8}{p{251pt}}{Please note that an object is always comprised of many frames, up to 300.}\\
    \multicolumn{8}{p{251pt}}{$^{\mathrm{a}}$B $=$ Background, J $=$ Jellyfish, A $=$ Artefacts, F $=$ Fish, Sw $=$ Seaweed, Sd $=$ Sediment.}
    \end{tabular}
\end{table}

\section{Improving Performance}
\subsection{Baseline System}
\begin{figure*}[!t]
\centerline{\includegraphics[width=.9\textwidth]{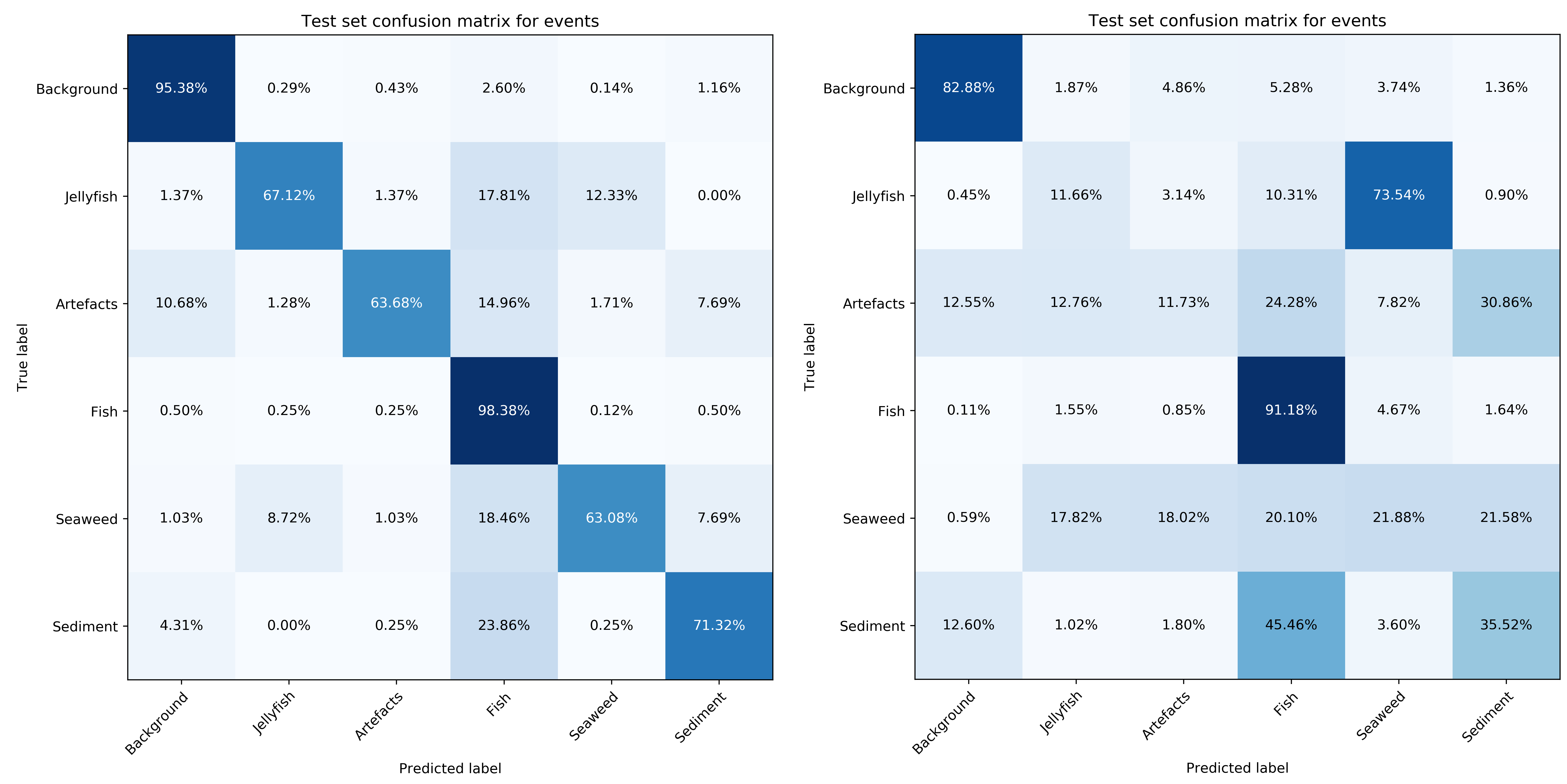}}
\caption{Confusion matrices illustrating the results of training and testing the system by drawing random splits from all the available imagery (left); training on Years 2015-2018 and testing on Year 2019 (right). The second approach shows how CNN fails to generalise on a number of classes apart from the 2 most popular ones: Background and Fish.}
\label{approach_comparison}
\end{figure*}
Before series of images are fed into a CNN classifier and predictions are produced, frames go through coordinate system conversion, noise reduction, object detection and object tracking steps. However, this paper focuses solely on improving classification results and, therefore, does not propose any modifications to the other parts of the system. A more detailed description of the pre-processing pipeline can be found in the previous work \cite{8652268}. Earlier experiments did not consider leave-one-out cross validation on different folds of acquired datasets (Table \ref{datasummary}), hence, the reported results did not prove that the model could generalise well, as both train and test sets were drawn from the same underwater environments. In other words, all imagery was put together and then split into 75\%/12.5\%/12.5\% chunks for training, validation and testing, respectively. In real life scenarios, test data will always come from a slightly different environment, with varying levels of noise and artefacts. Therefore, we change our evaluation strategy from randomly drawing splits from all the available footage to training and testing on separately acquired datasets. Fig. \ref{approach_comparison} shows the importance of testing on a dataset that is not present in the training split. Throughout this paper, we present several modifications that are aimed to improve classification results. To show their effectiveness, we report performance on the recently acquired data, Year 2019. It has the highest number of labelled objects, contains 38\% of all spotted jellyfish, and was collected in real life conditions (Table \ref{datasummary}). All reported results are calculated as an average over 10 randomly initialised runs. Table \ref{method_comparison} shows results achieved by various classification designs proposed in this paper.



\begin{table}
\caption{Results for various proposed classification frameworks}
\setlength{\tabcolsep}{4pt}
\centering
\label{method_comparison}
    \begin{tabular}{|p{90pt}|c|c|c|c|}
    \hline
         Method & Frame & Event & Jellyfish & Jellyfish FP \\ \hline \hline
        \texttt{A} Baseline system \cite{8652268} & 74.21\% & 69.80\% & 11.52\% & 2.26\% \\ 
        \texttt{B} + Generated data & \textbf{75.05\%} & 68.75\% & 10.85\% & 1.48\%\\
        \texttt{C} + Event classifier & 75.05\% & \textbf{69.88\%} & 18.07\% & 1.74\%\\
        \texttt{D} + Weighted loss & 75.05\% & 69.87\%  & \textbf{38.51\%} & 2.51\%\\
        \texttt{E} + Confidence threshold & 75.05\% & 69.87\% & 30.16\% & \textbf{0.91\%}\\
    \hline
    \multicolumn{5}{p{245pt}}{For frames, events and jellyfish, we report mean accuracy (higher is better). For jellyfish false positives (FP), we report the proportion of objects that are mislcassified as jellyfish (lower is better).}
    \end{tabular}
\end{table}

\subsection{Lack of Data}\label{Lack of Data}
Footage collection and annotation is an expensive and tedious process. Moreover, the objects of interest, such as jellyfish and seaweed, are not guaranteed to be present, which further amplifies the difficulty of acquiring desired imagery and results in an imbalanced dataset. Since deep neural network classifiers benefit from large amounts of samples, as that leads to better generalisation, data generation becomes one of the solutions. We implement state-of-the-art generative adversarial network - StyleGAN \cite{karras2019style} to supply synthetic samples in order to expand the training dataset. The StyleGAN framework is composed of two models: the Generator and  Discriminator, which are trained together. The purpose of Discriminator is to assign correct labels to both real and generated samples, while Generator is trained to minimise correct predictions by the Discriminator. The networks take turns in a minimax game, where improving the performance of one network negatively impacts the performance of the other. This leads to the non-saturating loss:
\begin{gather}
    L_D = log(D(x)) + log(1 - D(G(z))), \nonumber\\
    L_G = -log(D(G(z)))
\end{gather}
where $D$ is Discriminator, $G$ is Generator, and $z$ is random noise. Generated data $G(z)$ is labelled as 0, and real data $x$ is labelled as 1. Once trained, Discriminator is discarded, and Generator acts as a transform function which turns a fixed-length random vector into a sample that matches the desired distribution. An important extension to this framework is conditioning \cite{mirza2014conditional} - inputs to both models are accompanied with class labels. This way, the networks learn specifics of each category, and later can be used to generate samples of a given class. 
\begin{figure}[!t]
\centerline{\includegraphics[width=.8\columnwidth]{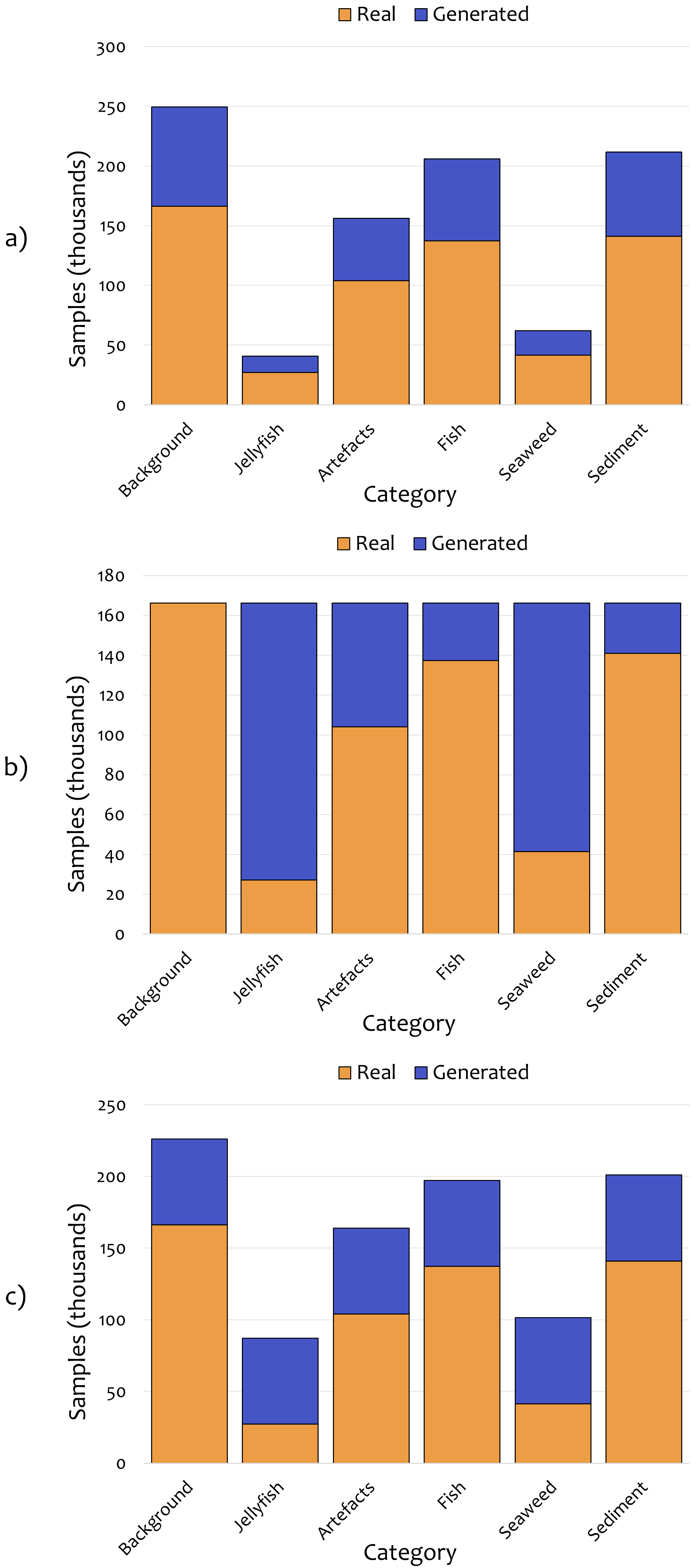}}
\caption{The chosen strategies for enhancing the train set with generated data: a) proportional to real data, b) proportional to missing data, c) adding the same number of samples to each class.}
\label{strategies}
\end{figure}

We leave models' architectures, hyperparameters and the training process unchanged, and train the networks until Discriminator has seen 25 million real images. We do not use the style-mixing feature proposed in the original paper \cite{karras2019style} to generate samples, though we leave it on during the training to enhance the quality of produced images. Both real and generated visual examples for each class can be found in the Supplementary Material \footnote{\url{https://docdro.id/etpDvBS}}. After being trained on Years 2015-2018, the Generator is used to produce 600 000 samples in total, 100 000 per category. To inspect the benefits of training on fake data in detail and find the most optimal strategy, it was decided to try 3 different enhancement strategies, shown in Fig. \ref{strategies}. Adding samples proportionally to the real ones (Fig. \ref{strategies}a) retains the same ratios between classes, which has very little chance of introducing data bias and, hence, increasing the number of false positives and false negatives. The second approach (Fig. \ref{strategies}b) is designed to fill the gaps and allow classes, such as Jellyfish and Seaweed, to have more impact on the loss which might force the CNN to learn visual features related to those categories better. Finally, adding the same number of images to each class (Fig. \ref{strategies}c) blends the preceding strategies together - it has little impact on category balance yet allows to add reasonable portions of samples to each.

We experimented with adding 10\%, 20\% and 50\% of fake samples with respect to the real data. For each setup, we report mean frame and event accuracy to show overall performance, as well as the sum of standard deviations over true positives to capture variance in predictions. We calculate the latter as follows: 
\begin{gather}
    y = \sum_{c=1}^{k}\sqrt{\frac{\sum_{i=1}^{n}(c_{x_i} - \overline{c_x})^2}{n-1}}
\end{gather}
where $k$ is the total amount of classes, $n$ is the total amount of confusion matrices and $c_{x_i}$ is the number of true positives for a particular run. For better readability, we normalise these numbers by dividing them by the value obtained for the first configuration. The results are shown in Table \ref{generated_results}.
\begin{table}
\caption{Mean frame and event accuracy and sum of standard deviations for frames and events for various data enhancement setups}
\setlength{\tabcolsep}{4pt}
\label{generated_results}
    \begin{tabular}{|l|c|c|c|c|}
    \hline
         Strategy & Frame & Event & Frame std & Event std \\ \hline \hline
         Only real data & 74.21\%$\pm$1.30 & \textbf{69.80\%$\pm$1.38} & \textbf{1.00} & \textbf{1.00} \\ \hline
        \texttt{a)} at 10\% & 74.19\%$\pm$1.58 & 68.00\%$\pm$1.05 & 1.12 & 1.13 \\
        \texttt{a)} at 20\% & 73.60\%$\pm$1.01 & 67.22\%$\pm$1.08 & 1.14 & 1.16 \\
        \texttt{a)} at 50\% & 74.32\%$\pm$1.93 & 67.68\%$\pm$2.15 & 1.38 & 1.51 \\ \hline
        \texttt{b)} at 10\% & 74.32\%$\pm$1.74 & 68.39\%$\pm$1.47 & 1.28 & 1.36 \\
        \texttt{b)} at 20\% & 72.35\%$\pm$1.99 & 66.79\%$\pm$2.18 & 1.36 & 1.38\\
        \texttt{b)} at 50\% & 71.21\%$\pm$2.28 & 66.36\%$\pm$1.82 & 1.87 & 1.89\\  \hline
        \texttt{c)} at 10\% & \textbf{75.05\%$\pm$1.22} & 68.75\%$\pm$1.08 & 1.10 & 1.09\\
        \texttt{c)} at 20\% & 73.49\%$\pm$3.16 & 68.21\%$\pm$1.95 & 1.36 & 1.36 \\
        \texttt{c)} at 50\% & 71.94\%$\pm$2.53 & 65.86\%$\pm$2.22 & 1.93 & 1.85\\
    \hline
    \multicolumn{5}{p{245pt}}{The referred strategies are shown on Fig. \ref{strategies}.}
    \end{tabular}
\end{table}

Clearly, adding synthetic data in small portions has the potential of slightly improving the system's performance. However, as the amount of fake images increases, a rapid growth in variance can be observed. We hypothesize that this behaviour could be caused by two factors:
\begin{enumerate}
    \item The generative model learned the training datasets too well. Therefore, the classifier becomes more biased when given additional imagery that lies within the same distributions. As a result, the model assigns incorrect labels to samples which are specific to that particular test set, e.g.: artefacts;
    \item The generative model produced too many noisy images which further complicates the classifier's training process.
\end{enumerate}
We stress that enhancing strategies are as important as the quality of fake data. The more the balance between classes is affected, the more variance in results can be observed, as well as drops in accuracy. Strategy \texttt{a)} kept ratios the same, and had the least negative impact in that regard. We chose the setting that has achieved the highest frame accuracy to work with in the subsequent experiments, i.e. strategy \texttt{c)} at 10\%. Whereas an improvement of 0.84\% in accuracy might seem not large enough, we stress that the ceiling is much lower than 100\% due to the human error in labels, as well as large amount of noise present in the data (e.g.: background frames are met in sequences, when the object is mostly faded but still being tracked). Some examples are presented in the Supplementary Material. Although event accuracy was not improved, we address this issue and focus on it in the next subsection.

\subsection{Event Information Fusion}\label{Event Information Fusion}
\begin{figure}[!t]
\centerline{\includegraphics[width=\columnwidth]{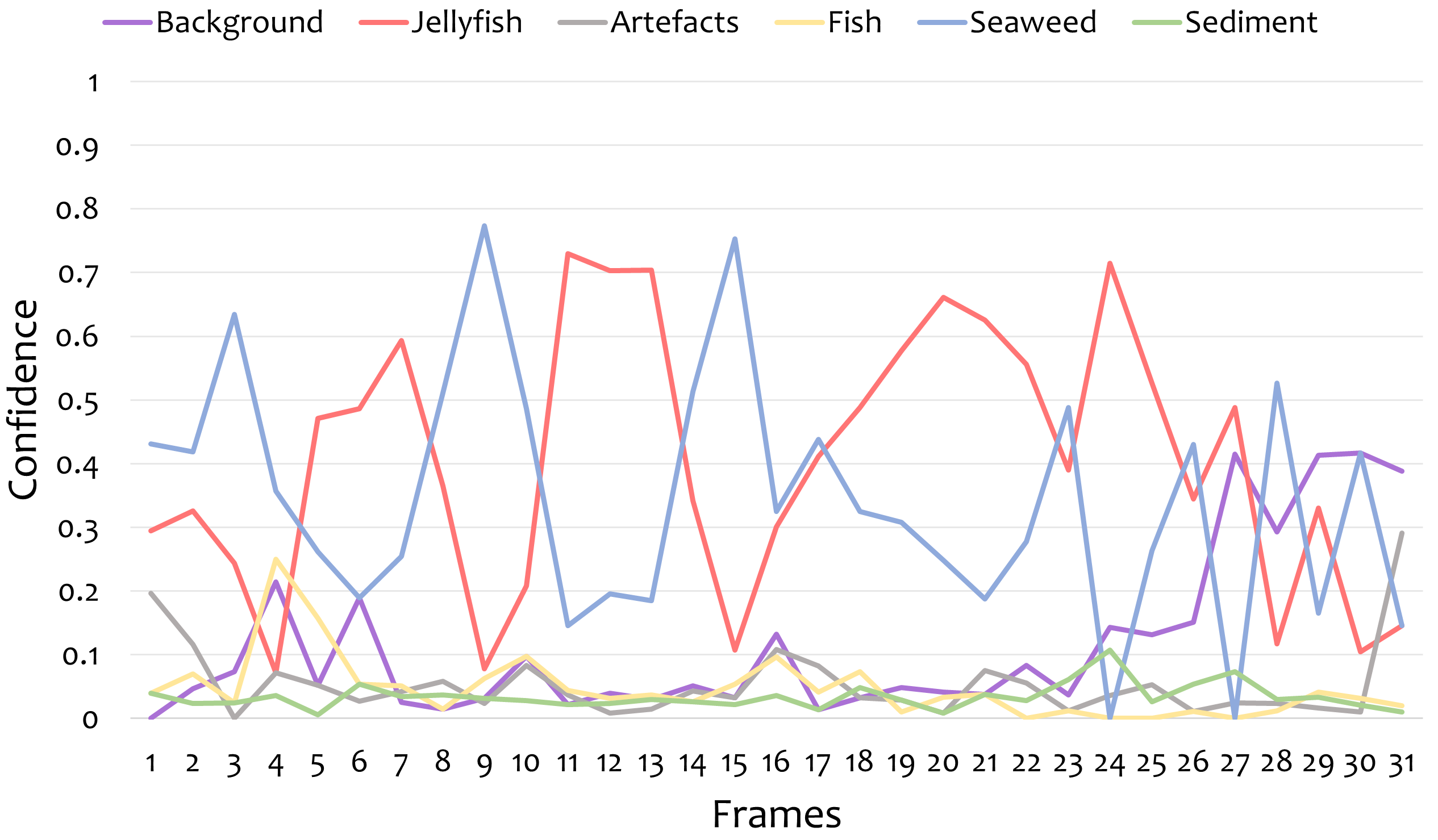}}
\caption{Confidence scores over a number of frames for a jellyfish. The pulsating motion causes a lot of variance in predictions - when the jellyfish contracts, it looks very similar to seaweed. Moreover, the first and last frames might often introduce additional noise, in cases when an object enters and/or exits the sonar's field of view, resulting in only a part of the object being visible.}
\label{confidence_pattern}
\end{figure}

\begin{figure}[!t]
\centerline{\includegraphics[width=\columnwidth]{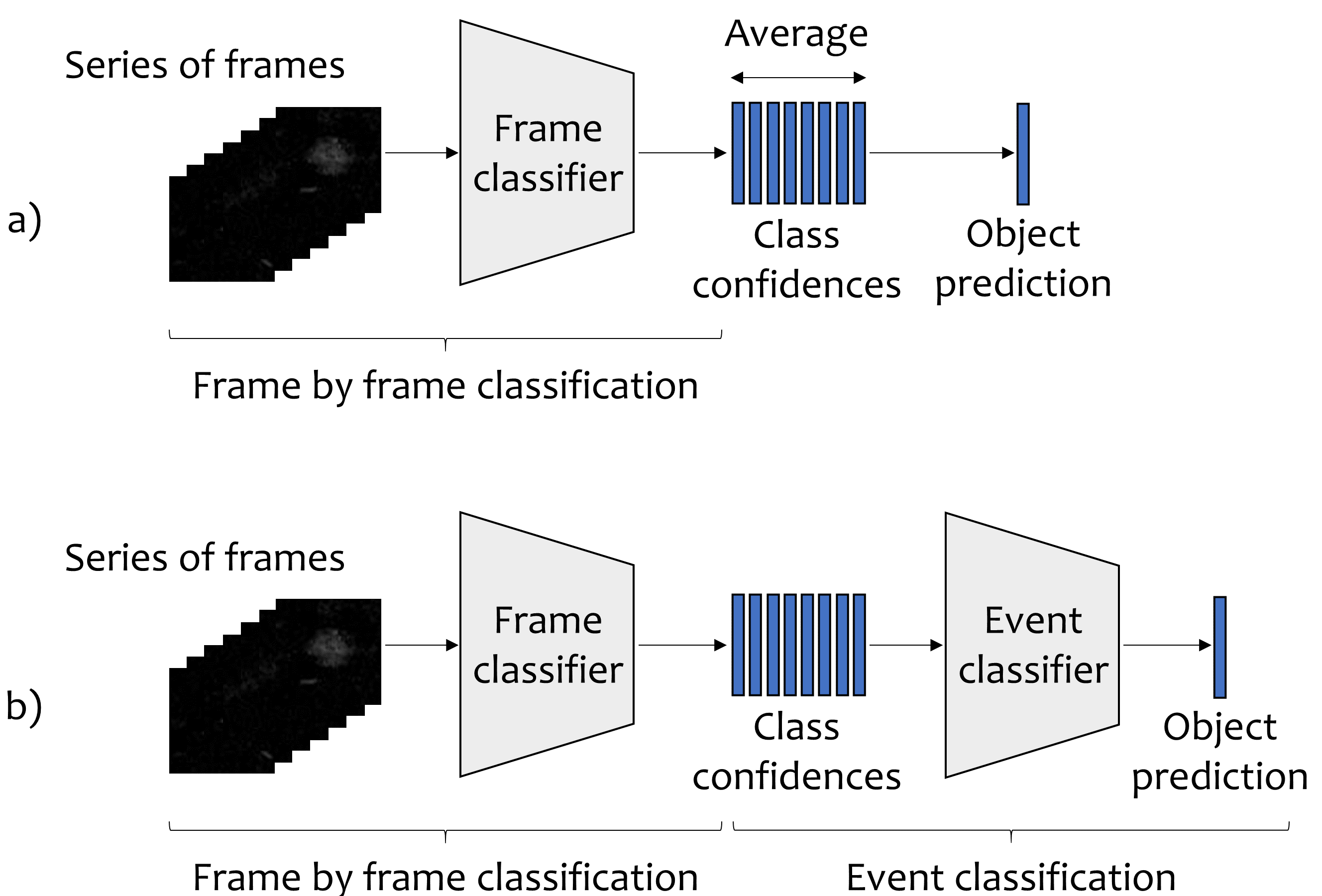}}
\caption{Visualization of two object classification frameworks: a) taking average over all individual frame predictions that belong to the same event to produce the final result \cite{8652268}; b) using an additional neural network to learn from confidence patterns, proposed in this paper.}
\label{event_cls}
\end{figure}

When the system stops tracking an object, a series of frames is extracted and passed to the CNN. Each event consists of a number of images ranging from 4 to 300, the average being 76. Frames are classified separately, then the average prediction is calculated to give a single output vector which contains class probabilities. Although this approach yields fairly good results, it relies on the idea that the majority of frames are classified correctly. However, often that is not the case - some objects might drastically change in shape and size. The problem is further amplified by changes in noise and artefact appearance. Deep neural networks are known to be very sensitive to even the slightest changes in data, and this results in incorrect predictions, which creates a gap between frame and event accuracies (Table \ref{method_comparison}, methods \texttt{A} and \texttt{B}). Fig. \ref{confidence_pattern} shows examples of rapid fluctuations in per-frame predictions. To combat this, one would have to come up with a method that considers objects' metadata instead of just taking the average. We experimented with changing the input to several images, adding mean and standard deviation over frames, and found that a network which convolves over confidence scores for series of frames in a sliding window manner works the best (Fig. \ref{event_cls}).

The motivation behind the proposed framework came from manual analysis of confidence patterns where we noticed that common features are often shared within one category, e.g.: small variance in shape for seaweed which leads to rather constant predictions over frames, pulsating motion for jellyfish that causes frequent changes, demonstrated in Fig. \ref{confidence_pattern}. If these patterns are consistently met across the datasets, the system can learn from them and reduce the confusion between certain classes as a result. Another benefit of the event classifier network is the small computational overhead, both in memory and time, which we analyze in detail at the end of this subsection. Table \ref{sliding_window} compares the two classification frameworks presented in Fig. \ref{event_cls}. To keep comparison fair, the same pre-trained frame classifiers were used when swapping techniques.

\begin{table}[]
\caption{Event accuracy for 10 randomly initialised runs using averaging and event classifier}
\setlength{\tabcolsep}{4pt}
\centering
\label{sliding_window}
    \begin{tabular}{|l|c|c|}
    \hline
        Run \# &  Averaging (Fig. \ref{event_cls}a) &  Event classifier (Fig. \ref{event_cls}b) \\ \hline \hline
        1 &  68.25\%  & 69.61\% \\
        2 &  67.43\%  & 69.85\% \\
        3 &  68.88\%  & 69.29\% \\
        4 &  68.47\%  & 69.44\% \\
        5 &  66.98\%  & 68.72\% \\
        6 &  68.72\%  & 69.10\% \\
        7 &  68.99\%  & 69.34\% \\
        8 &  69.18\%  & 71.50\% \\
        9 &  70.63\%  & 70.64\% \\
        10 &  69.96\% & 71.36\% \\
    \hline
    \end{tabular}
\end{table}

The suggested change outperformed averaging on every run by a margin of up to 2.42\%. However, the main purpose of JellyMonitor is to reliably recognize jellyfish without triggering any false alarms which can be caused by many false positives. Therefore, assessing the system's performance solely on frame and/or event accuracy is not the best approach in our application. Accuracy is mostly affected by categories with large amounts of samples - Background, Fish, and Sediment (Table \ref{datasummary}). Whereas it is important to keep this number high for monitoring purposes, a change that can bring the number of Jellyfish true positives up at the cost of some percentage of Sediment being classified as Background, for example, is considered a positive trade-off. That leads to the idea of a weighted loss - scaling the error up for some particular classes forces the model to assign correct labels more often when given samples from those categories. When reporting results, we also include Jellyfish accuracy and the number of false positives, which induce the risk of triggering alarms. The main confusion is between Jellyfish and Seaweed, as they happen to look similar on many frames. Moreover, these two are the least popular classes in the training dataset - only 546 Seaweed and 358 Jellyfish (Table \ref{datasummary}), mostly captured during trials, i.e. different to real underwater conditions. Hence, we experimented with scaling up the weights for them. It was impossible to make use of synthetic data to train the event classifier, as the generative network does not produce videos, only individual frames. Although we considered using weighted loss for the frame classifier too, the attempts did not lead to a desired accuracy increase, mostly due to the fact that the classifier is very sensitive to data imbalance. 

\begin{table}
\caption{Mean event and jellyfish accuracy and jellyfish false positives for different weighted loss setups}
\setlength{\tabcolsep}{4pt}
\label{weighted_loss}
    \begin{tabular}{|p{80pt}|c|c|c|}
    \hline
        $X, Y$ & Event & Jellyfish & Jellyfish FP \\ \hline \hline
        1.0, 1.0 & 69.88\%$\pm$0.95 & 18.07\%$\pm$7.44 & 282.1$\pm$143.7 \\
        1.5, 1.0 & \textbf{69.96\%$\pm$1.10} & 28.48\%$\pm$6.64 & 383.0$\pm$168.7 \\
        1.5, 1.5 & 69.85\%$\pm$1.05 & 20.36\%$\pm$6.29 & 315.7$\pm$147.7 \\
        1.5, 2.0 & 69.71\%$\pm$1.09 & 16.10\%$\pm$6.23 & 278.8$\pm$140.2 \\
        2.0, 1.0 & 69.87\%$\pm$0.95 & \textbf{38.51\%$\pm$9.85} & 408.6$\pm$168.2 \\
        2.0, 1.5 & 69.81\%$\pm$1.05 & 29.69\%$\pm$7.15 & 391.5$\pm$168.2 \\
        2.0, 2.0 & 69.58\%$\pm$1.01 & 21.75\%$\pm$7.18 & 380.1$\pm$237.0 \\
        2.0, 3.0 & 69.58\%$\pm$0.97 & 15.56\%$\pm$6.80 & \textbf{268.6$\pm$128.1} \\
        3.0, 2.0 & 69.58\%$\pm$0.99 & 32.60\%$\pm$7.49 & 423.3$\pm$175.0 \\
        3.0, 3.0 & 69.53\%$\pm$0.96 & 22.38\%$\pm$8.37 & 382.1$\pm$210.9 \\
    \hline
    \multicolumn{4}{p{245pt}}{$X$ and $Y$ are the weights for Jellyfish and Seaweed, respectively. All the other classes were set at 1.0. False positives range from 0 (best case) to 16250 (worst case).}
    \end{tabular}
\end{table}

Clearly, forcing the network to perform better on Jellyfish and Seaweed does not affect the overall event accuracy too much (Table \ref{weighted_loss}). There is an evident relationship between Jellyfish true and false positives - bringing Jellyfish accuracy up introduces more false detections, and vice versa. However, it is important that JellyMonitor is able to spot a bloom, hence, we choose the setting that has achieved the highest Jellyfish accuracy for our further experiments. We focus on eliminating false positives in subsection \ref{Confidence Threshold}.

\begin{figure}[!t]
\centerline{\includegraphics[width=.9\columnwidth]{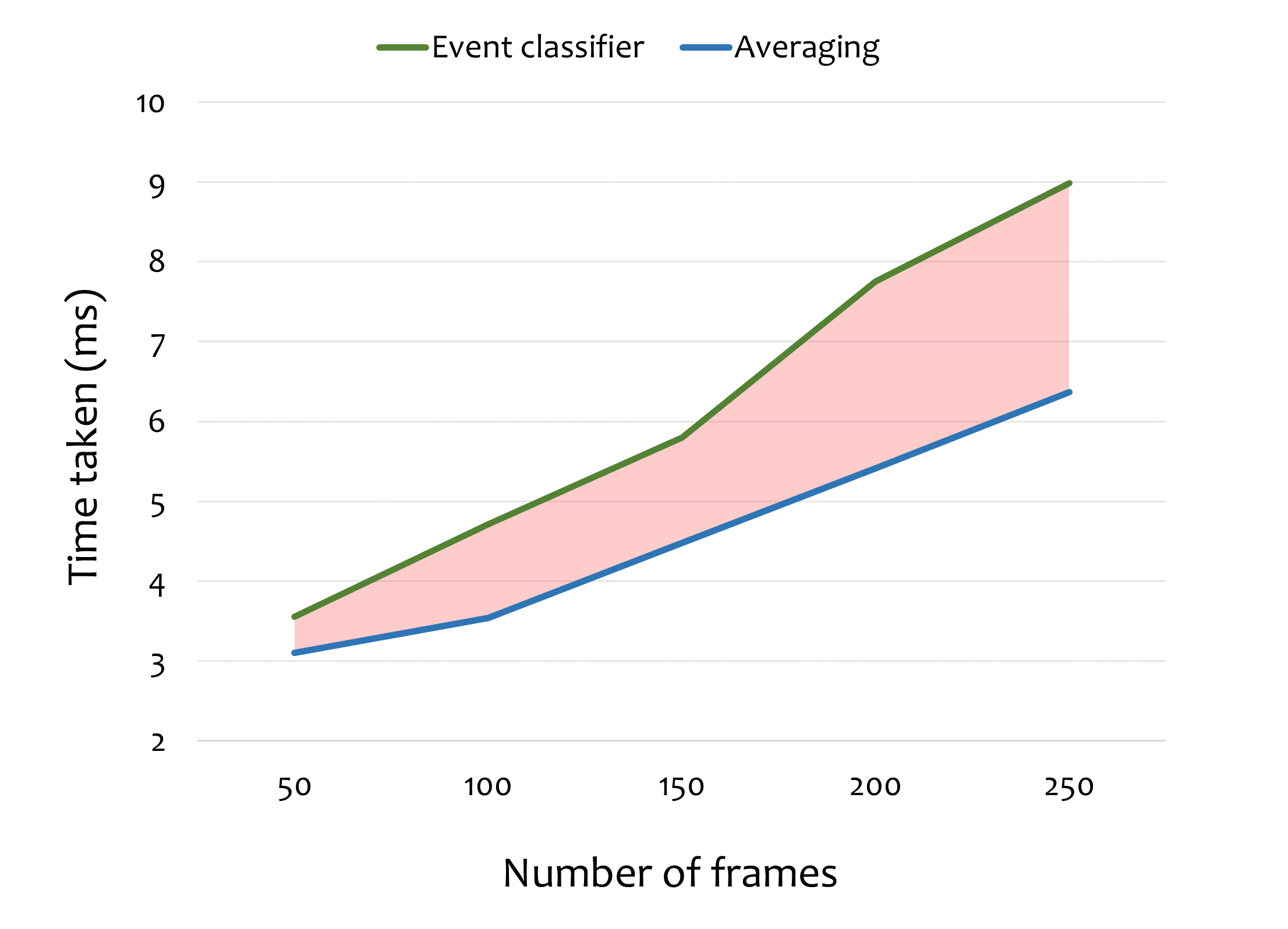}}
\caption{Average time taken to classify events with different frame lengths before and after adding the event classifier. Pink area shows the introduced time overhead.}
\label{time}
\end{figure}

It is crucial that the proposed technique does not introduce a lot of computational overhead - predictions should be performed in real time, and any major slow downs that could potentially lead to accumulation of data to be processed must be avoided. The event classifier performs well in that regard, slowing the system down by only up to 25\% for frame lengths of 150 and less (Fig. \ref{time}). Given that now footage capture happens only when the tidal flow is at its fastest, and objects move quickly, we do not expect many to surpass this threshold. As for GPU memory, it introduces additional 0.6 GB during test time with the current settings. When added to the initial 0.8 GB that come with the frame classifier, it is still way below 8 GB - the capacity offered by Nvidia Jetson TX2.

The proposed framework not only was able to provide a more structured approach to object classification, but also has given an opportunity to set priorities for classes with the help of weighted loss. That resulted in a significant boost of correctly classified jellyfish, however, introduced more false positives. Data collection strategy plays an important role - if before Year 2018 underwater phenomena moved freely most of the time, now they are almost always susceptible to the current, which affects the way objects change in shape over frames. Therefore, we believe that the event classifier will have much more positive impact during the next deployment, once Year 2019 is included in the training split.

\subsection{Confidence Threshold}\label{Confidence Threshold}
Objects that are assigned predictions with low confidence lie between major clusters on the classifier's decision surface. This is often an indication that samples are given wrong labels, and the probability of it being true increases as the amount of categories gets larger. It is important that JellyMonitor does not report false alarms - doing so would result in major losses for the industry. Therefore, we employ confidence thresholding, which allows to keep the amount of false positives low, while maintaining satisfactory true positive rate. Jellyfish is the only concern in this case, hence, we do not use thresholding for other classes.

\begin{figure}[!t]
\centerline{\includegraphics[width=\columnwidth]{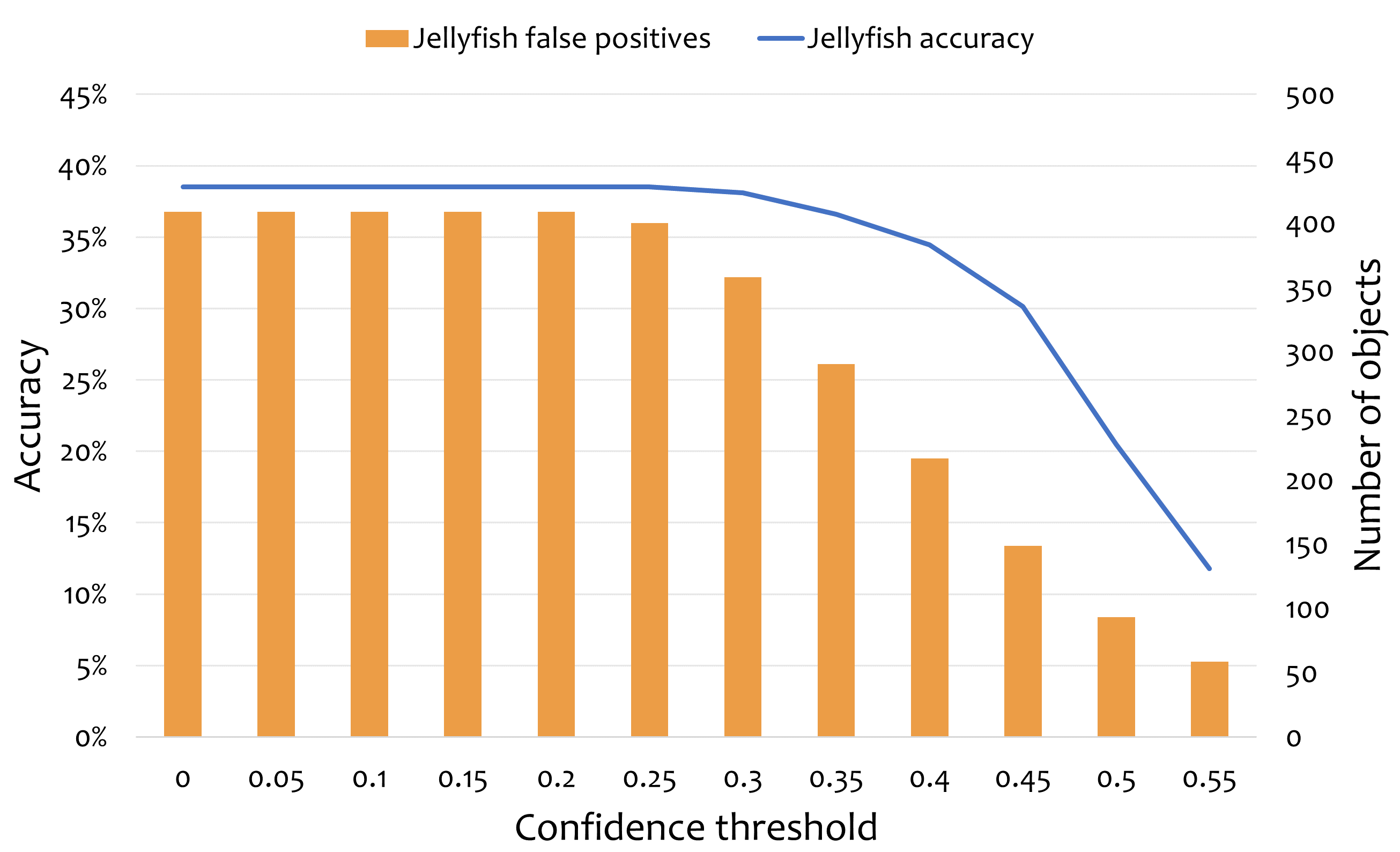}}
\caption{Average jellyfish accuracy and false positives for a series of threshold values.}
\label{threshold}
\end{figure}

\begin{figure}[!t]
\centerline{\includegraphics[width=\columnwidth]{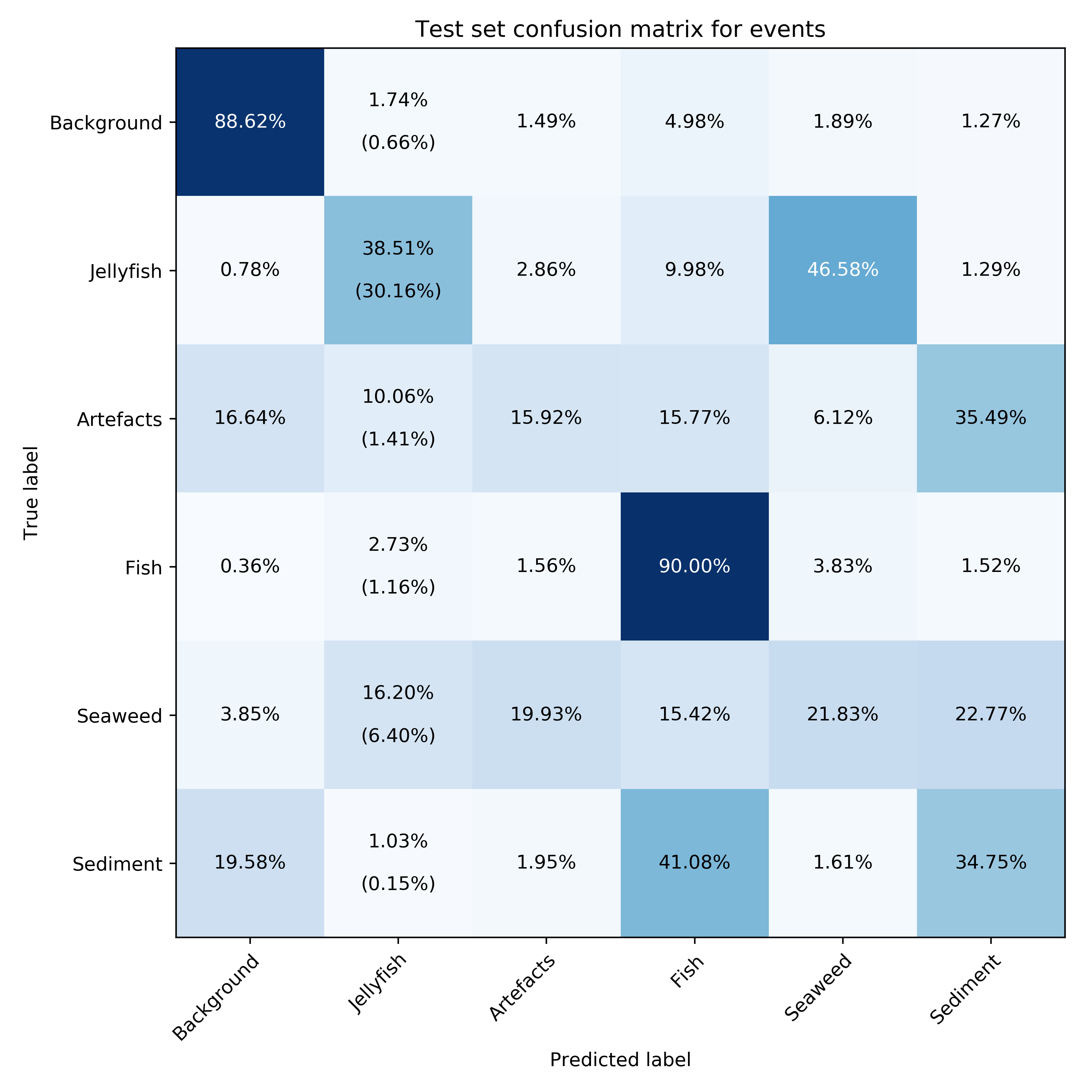}}
\caption{Final confusion matrix (average of 10 runs). Values in parentheses represent objects that surpass the threshold and are reported by the system to onshore.}
\label{final}
\end{figure}

The expected tradeoff can be observed between true and false positives (Fig. \ref{threshold}). To spot an abundance of jellyfish, it is enough to correctly classify a quarter - even with such a low accuracy, reported numbers will be high enough to indicate a bloom. Therefore, the chosen threshold value is 0.45, offering around 30\%, to accommodate for the possible error. The final confusion matrix is shown in Fig. \ref{final}. These are the results we expect the system to achieve in the next deployment, if not better. Events such as Background, Artefacts, Fish and Sediment will be sometimes misclassified as Jellyfish, however, even when combined together, these classes will not be mistaken as a bloom, due to extremely low error rate. Although Seaweed has a small chance of triggering an alarm, it would need to come in such an abundance that would cause the same negative effects on the industry as gelatinous plankton does. We would like to stress that the accuracy ceiling is much lower than 100\% due to the human error that was inevitably introduced during the labelling process, as well as large amount of noise present in the data (e.g.: background frames are met in sequences, when the object is mostly faded but still being tracked). 

\section{Conclusions and Future Work}
In this paper, we have presented a series of steps aimed at improving classification results for JellyMonitor \cite{8652268}. The proposed enhancements increased the jellyfish detection accuracy from 11.52\% to 30.16\%, and reduced the amount of false positives by more than 60\%. The system is now capable of spotting gelatinous zooplankton blooms with high confidence and can lower the damage and costs that jellyfish cause to a number of marine-related industries. Clearly, the system could be adapted to recognise other phenomena, and is not limited to the 6 mentioned categories, therefore, might be applicable to other underwater ecosystems and cases. 

The next stages of the project include further data collection and annotation, as well as deployments in coastal environments. We plan to continue employing deep learning methods, in particular, video data generation \cite{Tulyakov_2018_CVPR} to grow the dataset, transfer learning through domain adaptation to enhance the classifier \cite{DBLP:conf/icml/GaninL15}, and replacing object detection with more sophisticated techniques \cite{DBLP:journals/corr/abs-1804-02767}. The use of event metadata, such as motion paths, the speed of objects, the direction of the current, and the distance between the sonar and spotted phenomena can provide additional feature dimensions for event information fusion, and has great potential of improving results. 

\bibliographystyle{ieeetr}
\bibliography{jsen.bib}

\begin{IEEEbiography}[{\includegraphics[width=1in,height=1.25in,clip,keepaspectratio]{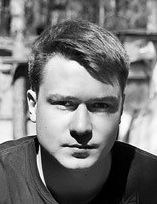}}]{Artjoms Gorpincenko} received the bachelor's degree in computing science from the University of East Anglia, Norwich, U.K., in 2018. He is currently pursuing the Ph.D. degree with the University of East Anglia. His research interests are related to computer vision and deep learning.
\end{IEEEbiography}

\begin{IEEEbiography}[{\includegraphics[width=1in,height=1.25in,clip,keepaspectratio]{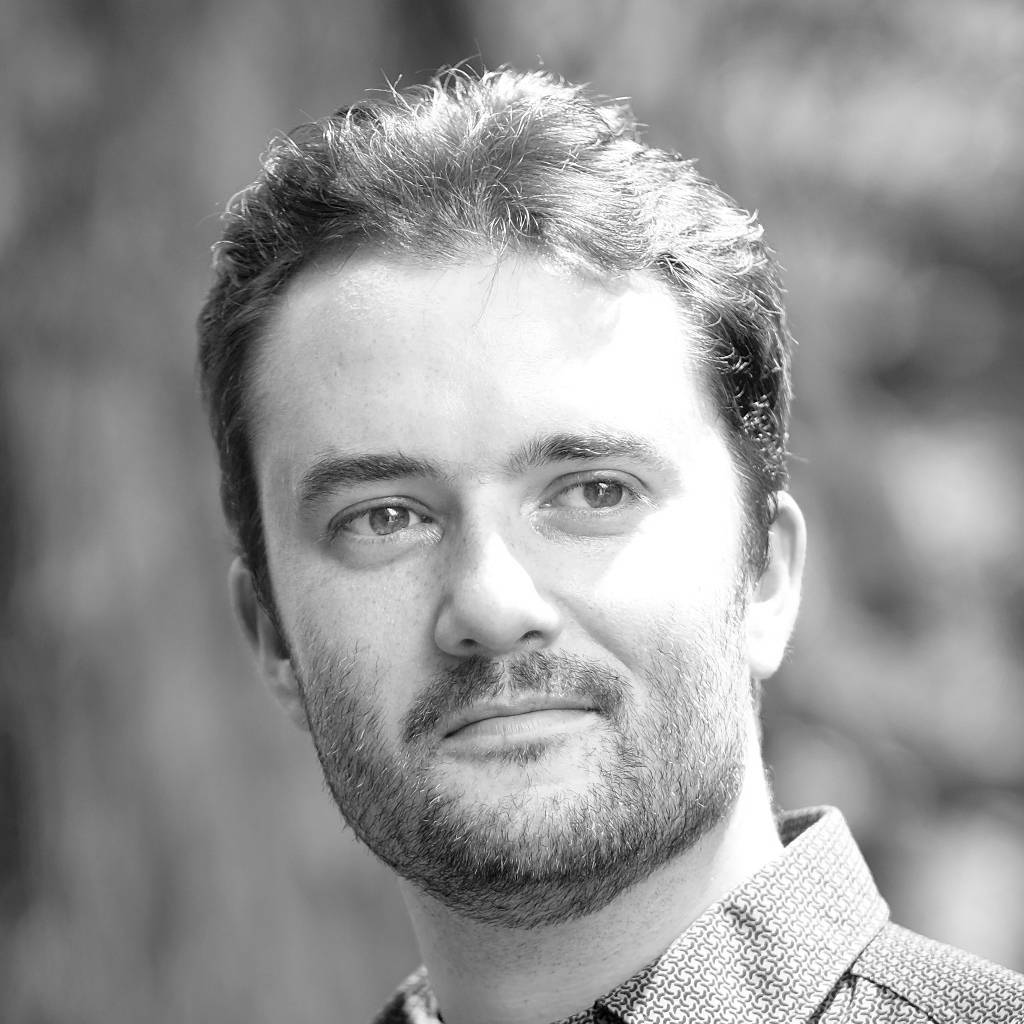}}]{Geoffrey French} received the bachelor's and master's degrees in computing science from the University of East Anglia, Norwich, U.K., in 2001 and 2013, respectively. e is currently pursuing the Ph.D. degree with the University of East Anglia. His research interests include computer vision and deep learning.
\end{IEEEbiography}

\begin{IEEEbiography}[{\includegraphics[width=0.01in,height=0.01in,clip,keepaspectratio]{a2.png}}]{Peter Knight} is currently working as an Electronic Engineer at Cefas Technology Ltd. Full biography and photo were not available at the time of publication.
\end{IEEEbiography}

\begin{IEEEbiography}[{\includegraphics[width=1in,height=1.25in,clip,keepaspectratio]{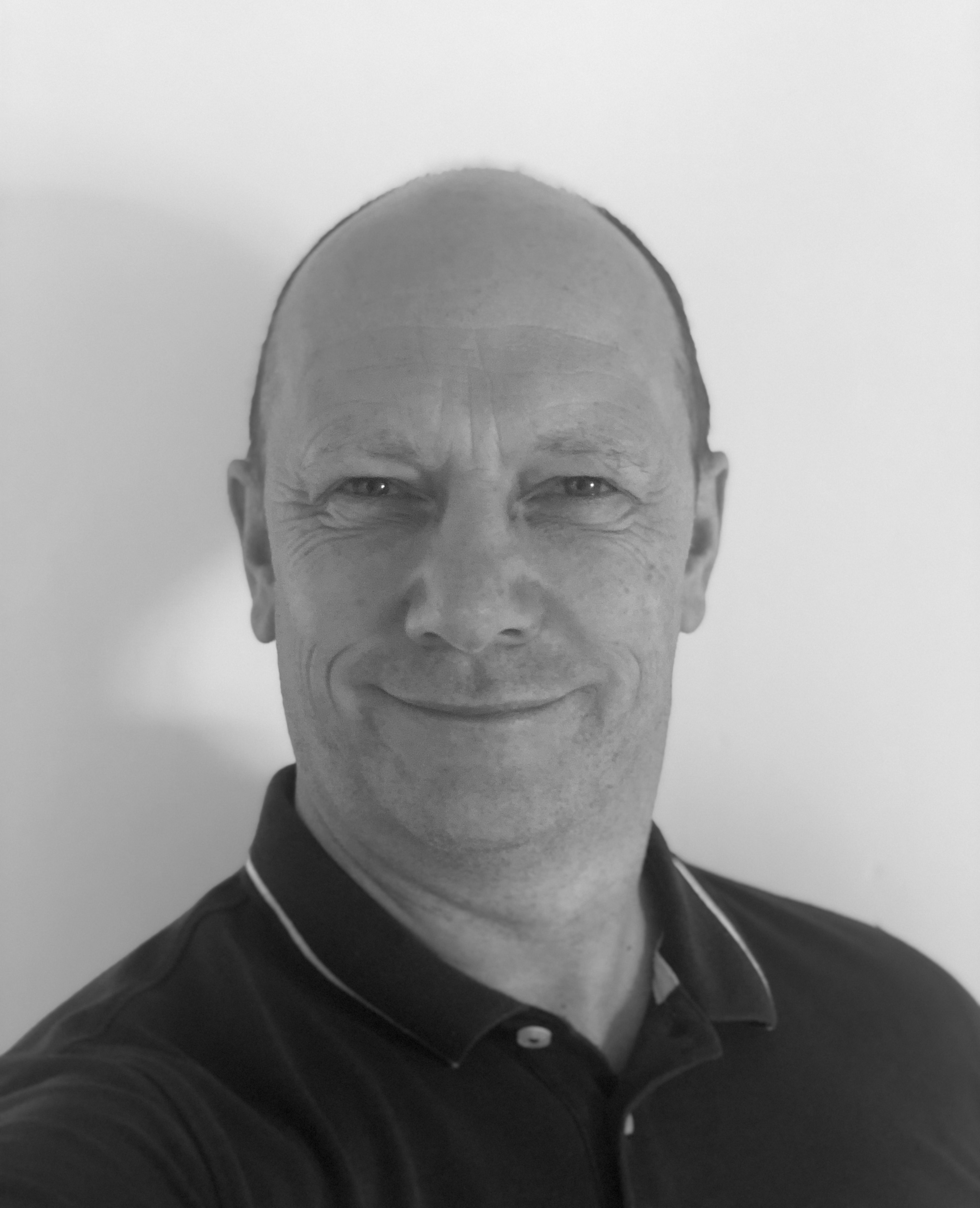}}]{Mike Challiss} is a Director of Cefas Technology Ltd, providing the development focus to technology projects for new markets. He is a senior project manager at Cefas and has managed key projects, such as our Research Vessel and more recently software technology projects with other UK Government departments. 

\end{IEEEbiography}

\begin{IEEEbiography}[{\includegraphics[width=1in,height=1.25in,clip,keepaspectratio]{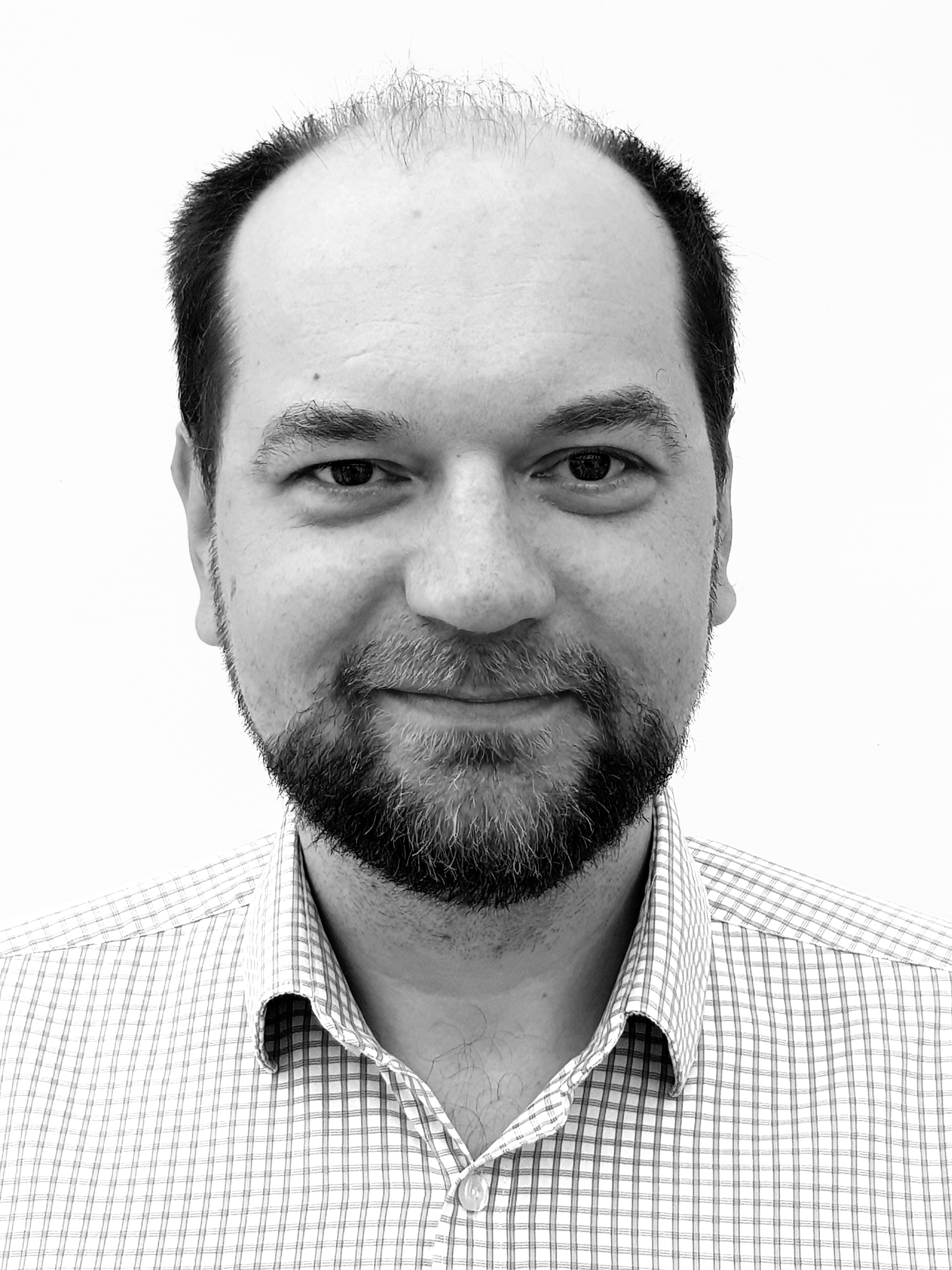}}]{Michal Mackiewicz} is an Associate Professor at the School of Computing Sciences, University of East Anglia (UEA). He received his MSc from the University of Science and Technology (AGH), Krakow, Poland in 2003 and the PhD from UEA in 2008. Michal has been involved in researching areas of colour science, physics based vision and machine learning. He has worked on a number of computer vision applications including medical imaging, remote sensing, environmental monitoring and agri-tech.
\end{IEEEbiography}

\end{document}